\pdfoutput=1
\documentclass[11pt]{article}
\usepackage[]{acl}
\usepackage{times}
\usepackage{latexsym}
\usepackage{graphicx}

\usepackage[T1]{fontenc}

\usepackage[utf8]{inputenc}

\usepackage{microtype}
\usepackage{float}
\usepackage{midfloat}
\usepackage{booktabs}

\title{Contextual effects of sentiment deployment \\ in human and machine translation}

  \author{%
  Lindy Comstock\textsuperscript{1,*}, Priyanshu Sharma\textsuperscript{2}, Mikhail Belov\textsuperscript{3}\\[1ex] 
   \footnotesize 1 Department of Psychiatry \&\ Biobehavioral Sciences, Semel Institute for Neuroscience and Human Behavior, \\ \footnotesize University of California, Los Angeles, Los Angeles, California, USA \\ 
    \footnotesize 2 Department of Linguistics, University of California, Los Angeles, Los Angeles, California, USA \\
  \footnotesize 3 Department of Communication, University of California, Los Angeles, Los Angeles, California, USA \\
  \footnotesize * corresponding author: \texttt{lbcomstock@ucla.edu}}

\begin{document}
\maketitle

\begin{abstract}

This paper illustrates how the overall sentiment of a text may be shifted in translation and the implications for automated sentiment analyses, particularly those that utilize machine translation and assess findings via semantic similarity metrics. While human and machine translation will produce more lemmas that fit the expected frequency of sentiment in the target language, only machine translation will also reduce the overall semantic field of the text, particularly in regard to words with epistemic content.

\end{abstract}

\section{Introduction}

Automated analyses are valued for their ability to quickly process huge quantities of text according to specific search parameters or research questions. At the same time, these methods are therefore at risk of overlooking the complex patterns inherent in human discourse and may instead treat language as a monolithic construct that remains invariant to genre-specific and contextual factors. Despite tremendous progress in the field \cite{min2023recent}, natural language processing (NLP) and deep learning techniques are still found to be lacking in the analysis of language pragmatics on the corpus level \cite{li2021semantics}, or may only achieve acceptable accuracy with the aid of human coding and annotation \cite{desmond2021semi,lebovitz2021ai}   

Automated approaches tend to result in an overgeneralization of language norms because their guidelines for text classification are distilled on a macro-level after training on huge datasets that span genres or integrate a range of texts thought to collectively represent a specific genre, such as political, media, or the types of discourse \cite{poria2014sentic}. This concept of language discounts the fact that pragmatic differences pervade even texts that are seemingly similar in nature, contingent on the speaker, speaker intent, environmental factors, and other contextual variables \cite{schegloff1999discourse,thomas2014meaning}. 

In this paper, we will discuss two scenarios that illustrate rarely-discussed sources of contextual variation which may lead to inaccurate findings in sentiment analyses performed on translated texts. The first requires an understanding of the normal distribution of sentiment in each language, and the second draws attention to widening and narrowing of the semantic field of a text which may not be apparent in the standard output metrics.   

\subsection{Background}

The existence of meaningful microscale variation in large language corpora has been successfully harnessed by scholars who conduct analyses on stylistics \cite{vishnubhotla-etal-2021-evaluation,zhu2022computational}, as well as those who utilize vector relationships to reveal unique sociolinguistic registers \cite{arseniev2020sociolinguistic}. While sociolinguistics has effectively used large-scale corpora to reveal contextual or group-level differences \cite{friginal2013corpus}, these analyses typically benefit from a clear understanding of the unit of analysis.

Phonetic or prosodic features will retain their meaning as a variable for search in ways that context-dependent affect or patterns of affect in a sentiment analysis may not. Moreover, many deep learning methods are "black box" models that elicit concerns about their interpretability and optimization because the feature sets and outputs cannot be equated with the traditional categorizations of human language \cite{zini2022explainability}. 

The challenges of classifying text corpora are compounded in crosscultural communication \cite{comstock2015facilitating}. Translation entails transformation of a text as it is refracted through the cultural experiences of the translator \cite{lotman1990universe}. In the absence of a structured and ubiquitous method for translation, each translator employs their own strategies, norms, and politics, which may distort the effect of the original text \cite{schaffner2004political} and the intended speaker intent \cite{araujo2020comparative}. 

We employ a transparent bag-of-words (BoW) sentiment analysis together with basic descriptive statistics to illustrate how a lack of understanding of the underlying properties of a corpus may produce misleading findings. While more sophisticated methods may generally be more accurate, we believe that discrepancies in the representation of underlying semantic relationships cannot fail to similarly affect the output of complex algorithms. 
    
\section{Contextual relevance}

We first report the sentiment classifications in a political source text and translation by (i) language, (ii) political context, and (iii) across presidential terms. Next, we compare expected and observed lemma frequencies for the Russian transcripts and their English translations. We found significant differences in all three categories, underscoring the linguistic and contextual specificity of how sentiment is deployed in real-world scenarios. 

\subsection{Method}

All the publicly available written transcripts of press conferences held by the Russian president at G8 and G20 summits between 2000-2015 were analyzed. The original \href{http://kremlin.ru}{Russian} transcripts and their official \href{http://en.kremlin.ru}{English} translations were accessed at the Kremlin online press archive. Data was taken from two international summits that represent maximally similar contexts. The Russian president was Vladimir Putin in 2000-2007 and 2013-2105 and Dmitry Medvedev from 2008-2011. The corpus is small in scope, yet it represents a comprehensive record of all the existing data from these summits (population-level metrics). It was selected because it has been extensively studied by the authors with automated and discourse analytic methods \cite{comstock2023journalistic,comstock2024sentiment}. 

\begin{table}[H]
\centering
\begin{tabular}{lccc}
\hline
\textbf{Summit} & \textbf{Term} & \textbf{Russian} & \textbf{English} \\
\hline
\verb|G8| & 2000-2003 & 757 & 874 \\
\verb|| & 2004-2007 & 2129 & 2611 \\
\verb|| & 2008-2011 & 1412 & 1709\\ 
\verb|| & 2012-2015 & 611 & 737 \\
\verb|G20| & 2000-2003 & -- & --\\
\verb|| & 2004-2007 & -- &  --\\ 
\verb|| & 2008-2011 & 1598 & 1887 \\
\verb|| & 2012-2015 & 2241 & 2474 \\\hline
\verb|Total| &  & 12338 & 14667 \\\hline
\end{tabular}
\caption{The total word count of the Russian transcripts and English translations for each political summit.}
\label{tab:accents}
\end{table}

Lists of positive, negative, and epistemic words were compiled from the Harvard IV-4, Loughran, McDonald, and Lexicoder sentiment dictionaries. Each lemma was restricted to one of three sentiment lists (positive, negative, subjective) to ensure that cross-listed words would not force a correspondence between sentiment classification results. The translation accuracy of the sentiment lists was confirmed by a professional Russian translator. 

We compiled (i) the number of lemmas in each language dataset, and calculated (ii) the observed frequency of each lemma per dataset as a percentage of the total words, and (iii) the expected frequency of each lemma as a percentage of its observed frequency in a larger corpus assumed to be representative of wider language norms. The expected frequencies were calculated using the sub-corpus of media texts from the \href{https://ruscorpora.ru/}{Russian National Corpus} and the \href{https://books.google.com/ngrams/}{English Google Ngrams corpus}. The total word counts, observed frequency, and expected frequencies were recorded for language, summit, and term. We performed an ANVOA and Tukey ASD test in JASP \cite{JASP2024} for each analysis. 

\subsection{Results}

\begin{figure*}[ht]
\centering
    \includegraphics[width=1\textwidth]{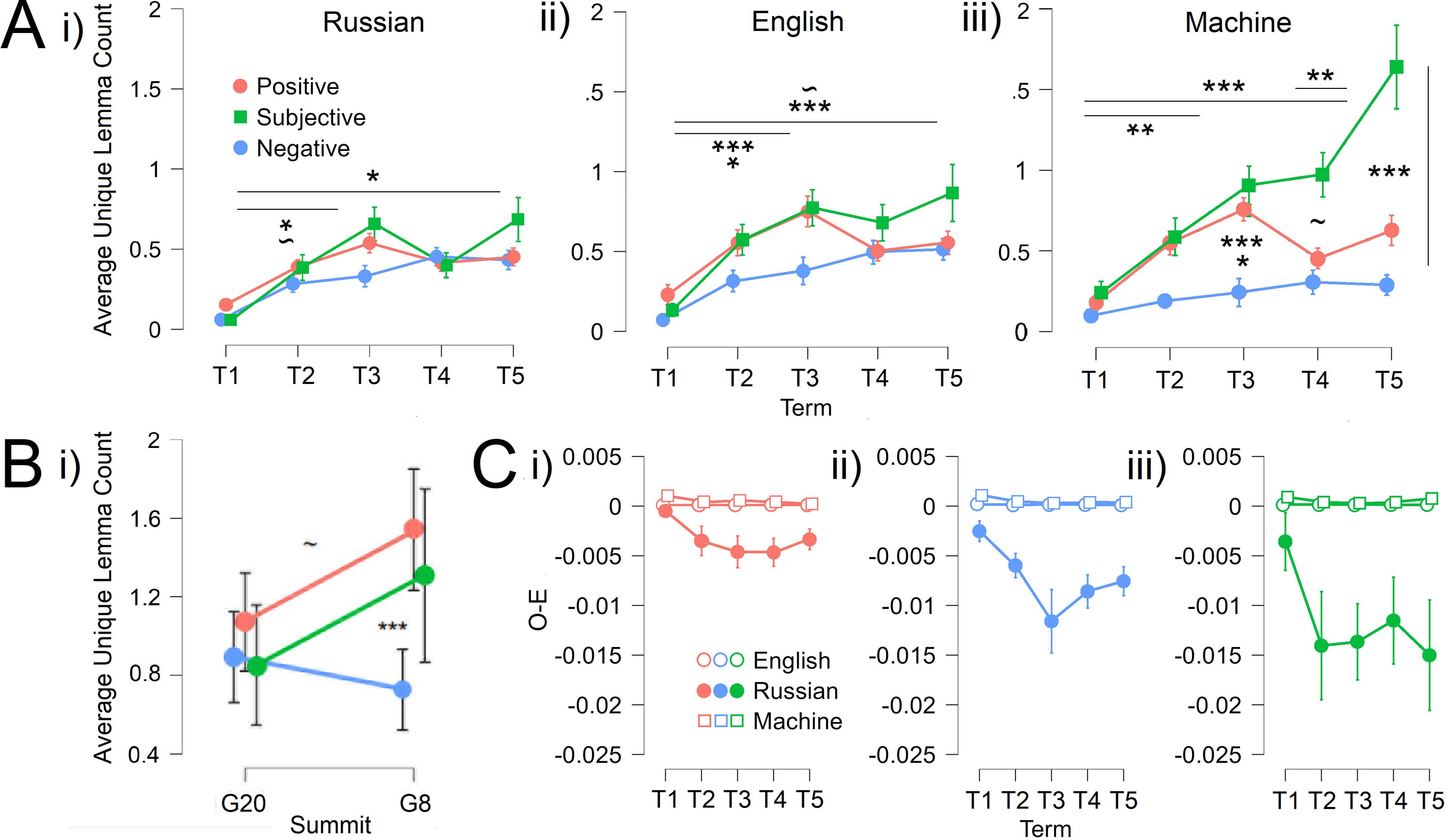}
\caption{\textbf{A.} The average unique lemma count per sentiment list over five presidential terms for (i) the source text and (ii) human and (iii) machine translation. \textbf{B.} The average unique lemma count per sentiment list across summits. \textbf{C.} Differences in the expected frequency of lemmas in a general language corpus from the observed frequency of lemmas in the combined summit dataset for \textbf{(i)} positive words, \textbf{(ii)} negative words, and \textbf{(iii)} epistemic words.}
\label{fig:SCiL}
\end{figure*}

Fig. \ref{fig:SCiL}A(i) presents the average count of the unique lemmas produced from each sentiment list across five presidential terms. The composition of words expressing sentiment does not remain consistent over even short time periods. Presidential terms here are not simply representative of time: previous research has shown that presidents are questioned differently depending on whether they are in their first term or returning to the office \cite{clayman2006historical}; as anticipated, in the original source text, the sentiment expressed by journalists questioning the Russian president most strongly differs from other terms when Medvedev first takes the office. 

Clear distinctions in the sentiment expressed by journalists in each context also appear in Fig. \ref{fig:SCiL}B(i): a greater number of unique positive and subjective lemmas are produced in the G8 summit than in the G20 summit. There is a marginally significant increase in positive lemmas between summits, and the G8 summit shows a significant difference between positive and negative lemmas. Although G8 data spans four terms and G20 data comprises only the last two terms, this difference cannot account for the finding: positive lemmas are at their highest level in the second two terms

The most striking effects appear when sentiment is compared in translations produced by a human (Fig. \ref{fig:SCiL}A(ii)), and the \href{https://translate.google.com/}{Google translate} website (Fig. \ref{fig:SCiL}A(iii)). The trends observed in the source text are augmented in human translation, resulting in greater change between terms. In the machine translation, the trends are exaggerated, particularly from the third term onward. The substantial rise in epistemic lemmas during the fifth term could even be considered a misrepresentation of the data.  

Fig. \ref{fig:SCiL}C(i-iii) illustrates the observed frequency of unique lemmas in summit data relative to their expected frequency in a general purpose corpus (\href{https://books.google.com/ngrams/graph?content}{Google Ngrams} or the \href{https://ruscorpora.ru/en}{Russian National Corpus}). Both human and machine translation match the lemma frequency observed in a general corpus, whereas the source text uses less emotive words than expected. Translation appears to misrepresent the underlying semantic field of the source text, instead reflecting target language expectations. Given the higher total word count found in the translated texts, in the absence of a real trend we might expect the reverse to occur: the translated lemma frequencies would be lower than expected compared to Russian, due to the larger English word count.

\begin{figure}
\centering
    \includegraphics[width=0.47\textwidth]{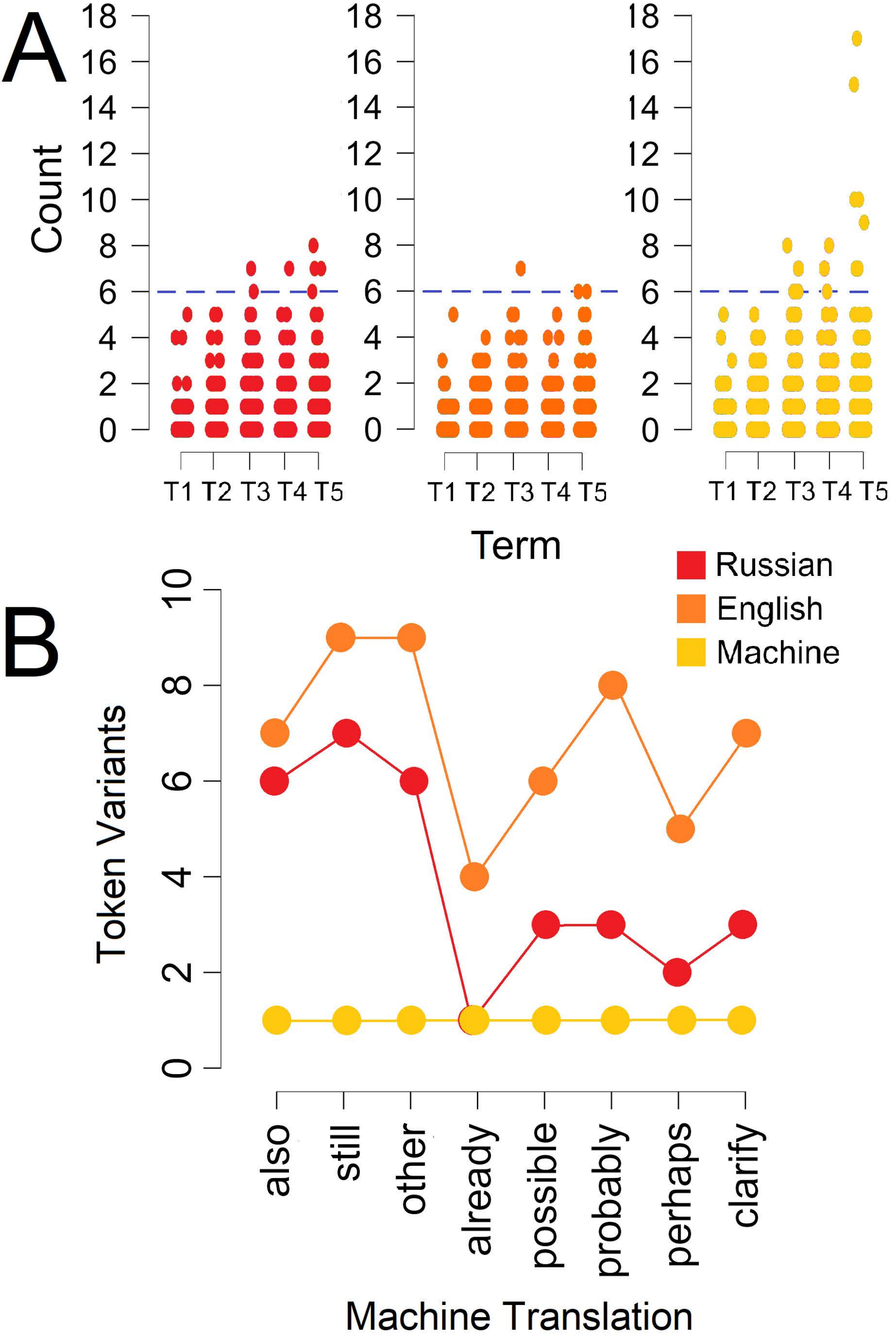}
\caption{\textbf{A.} Token counts per lemma. \textbf{B.} Translation variants per most frequent lemmas, across languages.}
\label{fig:Counts}
\end{figure}

\newpage
Notably, when a lemma with synonyms has many counts (tokens), fewer synonyms appear in the corpus. Fig. \ref{fig:Counts} illustrates the token number per lemma across languages. Human translation slightly increases the unique lemma count, thereby reducing the token count per lemma, whereas the machine translation increases tokens and reduces synonyms. Fig. \ref{fig:Counts}B(i)) plots the
synonyms used for a set of lemmas (those with the highest token count in machine translation). We observe a widening of the semantic field in human translation (> synonyms, < counts per lemma), as the translator embellishes the style of the original text, selecting the best synonym for each specific context. To the contrary, machine translation reduces the semantic field. These trends will be explored in a larger dataset to test these claims. 

\section{Semantic narrowing}

Our research questions were next investigated in the sentiment classification of literary source texts and their human and machine translations by (i) language, and (ii) author. We compare the total lemma and unique lemma counts by sentiment list. Similar trends to those discovered in the smaller political dataset were identified, providing further evidence that the deployment of sentiment in translations can deviate from the sentiment of the source text in predictable ways and may reflect a universal concern for sentiment analyses of all data types. 

\subsection{Method}

Three classic Russian novels served as source texts for analysis: \underline{The Double} \cite{dostoevsky2009double}, \underline{A Hero of Our Time} \cite{lermontov2019}, and \underline{The Captain's Daughter} \cite{pushkin2018captain}. Translations were generated via (i) machine translation on the Google translate website and (ii) scraping human translation from the opensource \href{https://gutenberg.org/}{Project Gutenberg} online archive. Each novel yields a dataset between 3-5 times larger than the summit corpus (a total corpus nearly 13 times larger).

\begin{table}[H]
\centering
\begin{tabular}{lccc}
\hline
\textbf{Author} & \textbf{Russian} & \textbf{Machine} & \textbf{Human} \\
\hline
 \verb|Dostoevsky| & 61589 & 1219 & 1459 \\
  \verb|Lermontov| & 57841 & 1332 & 4453 \\
   \verb|Pushkin| & 38590 & 1105 & 1092 \\
   \hline
       \verb|Totals| & 158020 & 3896 & 3507 \\
          \hline
\end{tabular}
\caption{The total word count for (i) source texts, and the total unique lemma counts for sentiment words in (ii) machine translation, and (ii) human translation.}
\label{tab:accents}
\end{table}

To underscore how the information revealed by the standard reporting for sentiment analyses and semantic similarity metrics differs from our data, we first briefly present the vector embeddings for one literary text (Lermontov): (i) principal component analysis (PCA), (ii) cosine similarity, and (iii) Euclidean distance. These analyses were performed on the uncurated word lists, without initial segmentation of the data into sentiment lists.

\section{Results}

\subsection{PCA, cosine similarity, and Euclidean distance measures}

\begin{figure}[!h]
    \includegraphics[width=.45\textwidth]{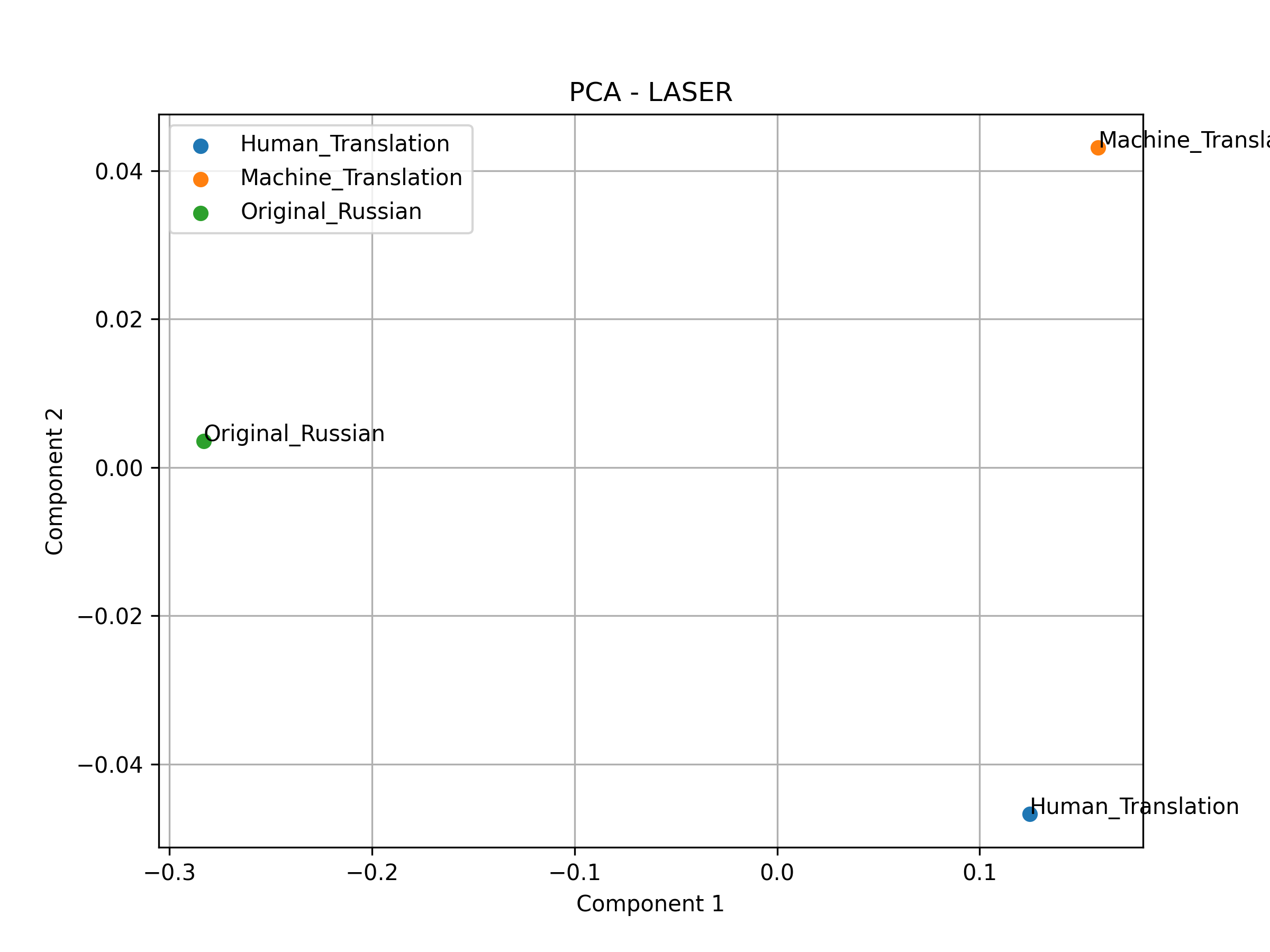}
\caption{PCA of the Russian source text and human and machine translations.}
\label{fig:PCA}
    \includegraphics[width=.5\textwidth]{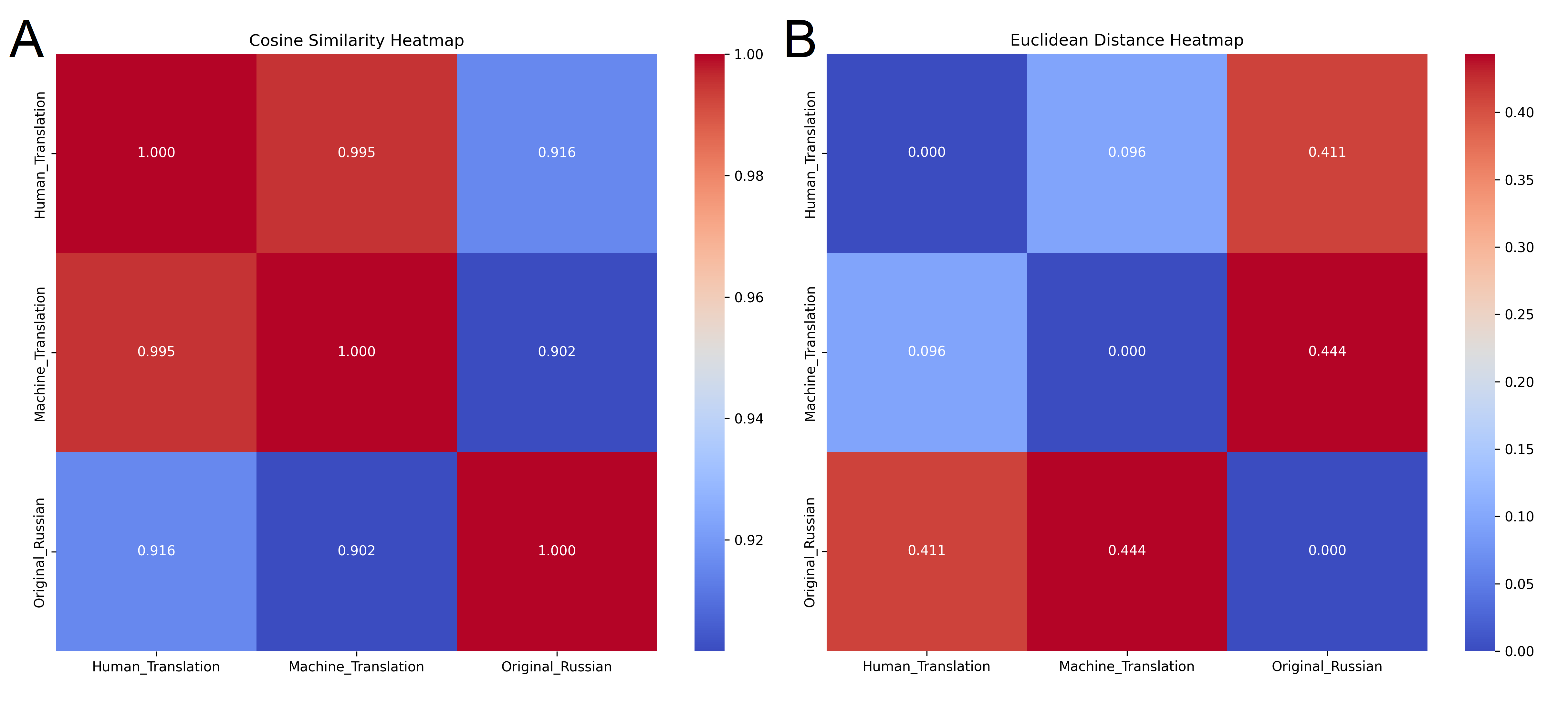}
\caption{The cosine similarity and Euclidean distance between the Russian source text and the human and machine translations.}
\label{fig:Cosine}
\end{figure}

PCA (LASER) graphically visualizes the distance and direction between vectors within a multi-dimensional embedding space (Fig. \ref{fig:PCA}) according to two components, whereas cosine similarity metrics measure the cosine of the angle between the two vectors of interest (Fig. \ref{fig:Cosine}A). Euclidean distance computes the straight-line distance between the points represented by the vectors (Fig. \ref{fig:Cosine}B). While the visualizations reveal human translation is more semantically similar to the source text and the translations are hugely similar, the features driving these differing assessments are not revealed. 

\begin{figure*}[ht]
\centering
    \includegraphics[width=.97\textwidth]{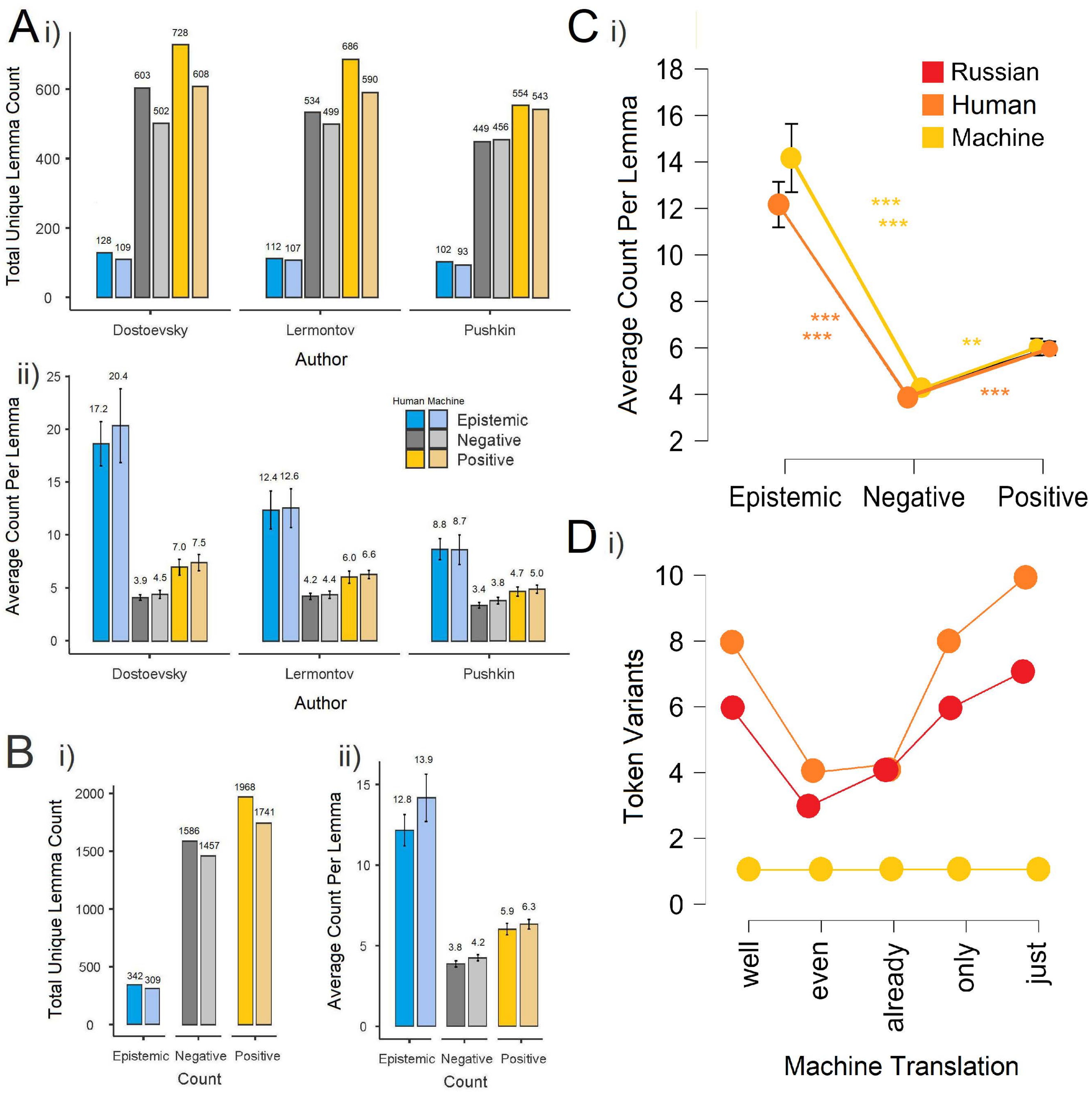}
\caption{Unique lemmas by author and sentiment. \textbf{A.} The total number of unique lemmas. \textbf{B.} The average count (tokens) per lemma. \textbf{C.} The average count per lemma in a combined corpus with significance levels. \textbf{D.} Translation variants per most frequent lemmas, across languages.}
\label{fig:Lit}
\end{figure*}

\subsection{BoW sentiment analysis}

Similar findings were observed across literary texts with variation that corresponds to known differences in total word count per novel. Human translation consistently produced a larger number of unique lemmas for each sentiment list (Fig. \ref{fig:Lit}A(i)). Machine translation generated fewer unique lemmas, and thus they were utilized more often, resulting in a slightly larger average count per lemma (Fig.  \ref{fig:Lit}A(ii)). The similar findings suggest that we can combine the datasets into a larger corpus to assess the observed trends statistically. We observe a statistically significant difference between translation types for epistemic lemmas (Fig. \ref{fig:Lit}C(i)).

Additional notable finding concern the role of epistemic lemmas: although considerably fewer in number, their usage remains high, irrespective of total word count. Epistemic lemmas therefore appear to be one factor that shapes authorial style and the semantic fields of the source text. Just as in the political database, we observe that machine translation narrows the semantic field for epistemic content. The five most frequent epistemic lemmas produced by machine translation are presented in (Fig. \ref{fig:Lit}D(i)). The source text produces multiple synonyms for the same set of lemmas, and the human translation even expands the semantic field slightly beyond that of the original novel. 

\section{Discussion}

Our findings illustrate several key differences that can inform sentiment analyses. Firstly, pragmatic norms in the assignment of emotional and subjective words vary by a variety of contextual factors, as well as between a source text and translation. These differences are exacerbated between human-generated and machine-generated translation, with human translation providing better representation of the underlying sentiment expressed in the source text. Secondly, in both cases, the observed frequencies of sentiment levels (positive, negative, and subjective lemmas) in a translation may be transformed to correspond to the expected sentiment levels calculated from general language sources; the observed frequencies of sentiment levels differ significantly from expected general language norms in the source text. Thirdly, polysemous or semantically amorphous words (typically subjective words) appear to be particularly prone to distortion in the translation process. They are provided with a greater contextual specificity in human translation, and reduced specificity in machine translation. Finally, common metrics to asses the semantic similarity between a source text and translation fail to register the widening or narrowing of the semantic field described here. 

\section{Conclusion}

Overall, translation normalizes pragmatic strategies, deploying sentiment according to language-specific expectations. This phenomenon occurs irrespective of the source or size of the corpus. Machine translation normalizes pragmatic strategies by considerably reducing the semantic field by selecting the most common translation without multiple synonyms, while human translation does so by somewhat expanding the semantic field in a contextually-relevant fashion. We propose that the underlying vector relationships cannot fail to be transformed given a manipulation of the scope of the semantic field and sentiment assignments. Therefore, the topic warrants further discussion in the literature.

\bibliography{anthology,main}

\bibliographystyle{acl_natbib}
\end{document}